\documentclass[journal]{IEEEtran}
\usepackage{graphicx}
\usepackage{amsmath}
\usepackage{multirow}
\usepackage{fancyvrb}
\usepackage{xcolor}

\usepackage[ruled, lined, longend, linesnumbered]{algorithm2e}

\usepackage{algorithmic}
\usepackage{environ}
\usepackage{tikz}

\NewEnviron{elaboration_red}{
\par
\begin{tikzpicture}
\node[rectangle,minimum width=0.37\textwidth] (m) {\begin{minipage}{0.35\textwidth}\BODY\end{minipage}};
\draw[dashed,color=red] (m.south west) rectangle (m.north east);
\end{tikzpicture}
}

\NewEnviron{elaboration_green}{
\par
\begin{tikzpicture}
\node[rectangle,minimum width=0.37\textwidth] (m) {\begin{minipage}{0.35\textwidth}\BODY\end{minipage}};
\draw[dashed,color=green] (m.south west) rectangle (m.north east);
\end{tikzpicture}
}

\ifCLASSINFOpdf
\else
\fi
\begin{document}
\title{Learning Pairwise Relationship for Multi-object Detection in Crowded Scenes}

\author{Yu Liu, Lingqiao Liu, Hamid Rezatofighi, Thanh-Toan Do, Qinfeng Shi and Ian Reid% <-this % stops a space
       
% \thanks{Manuscript received xx xx, xxxx; revised xx xx, xxxx.}

\thanks{Y.~Liu, L.~Liu, H.~Rezatofighi, Q.~Shi and I.~Reid are with the School of Computer Science, the University of Adelaide, Adelaide, SA 5005, Australia (e-mail: yu.liu04@adelaide.edu.au; lingqiao.liu@adelaide.edu.au; hamid.rezatofighi@adelaide.edu.au; javen.shi@adelaide.edu.au; ian.reid@adelaide.edu.au;
). (Corresponding author: Yu Liu)}%
\thanks{T-T.~Do are with the Department of Computer Science, University of Liverpool, Liverpool, L69 3BX, UK (Thanh-Toan.Do@liverpool.ac.uk).}%
}


\markboth{
% IEEE Transactions on Circuits and Systems for Video Technology,~Vol.XX, No.XX, XX XXXX
}%
{Shell \MakeLowercase{\textit{et al.}}: Bare Demo of IEEEtran.cls for IEEE Journals}



\maketitle

\begin{abstract}
As the post-processing step for object detection, non-maximum suppression (GreedyNMS) is widely used in most of the detectors for many years. It is efficient and accurate for sparse scenes, but suffers an inevitable trade-off between precision and recall in crowded scenes. To overcome this drawback, we propose a Pairwise-NMS to cure GreedyNMS. Specifically, a pairwise-relationship network that is based on deep learning is learned to predict if two overlapping proposal boxes contain two objects or zero/one object, which can handle multiple overlapping objects effectively. Through neatly coupling with GreedyNMS without losing efficiency, consistent improvements have been achieved in heavily occluded datasets including MOT15, TUD-Crossing and PETS. In addition, Pairwise-NMS can be integrated into any learning based detectors (Both of Faster-RCNN and DPM detectors are tested in this paper), thus building a bridge between GreedyNMS and end-to-end learning detectors.
\end{abstract}

\begin{IEEEkeywords}
Multi-object detection, Pairwise-learning, Greedy-NMS.
\end{IEEEkeywords}

\IEEEpeerreviewmaketitle

\section{Introduction}

\IEEEPARstart{M}{ulti-object}
%% 2.1 the problem of heavy overlap for multi-object detection
detection is a vital problem in computer vision, and can be viewed as the cornerstone for instance-level segmentation~\cite{Reference35,Reference11,Reference36} and multi-object tracking~\cite{Reference17,Reference29,Reference18}. In multi-object detection, several objects are usually overlapping with each other as a cluster, which is called co-occurrence~\cite{Reference30,Reference31,Reference32}. The problems that arise with co-occurrence are challenging such as occlusion, illumination changing, texture inconsistency, etc. Among these problems, occlusion is the most notorious issue. Based on whether the clusters of objects belong to the same category or not, the co-occurrence can further be classified into intra-class co-occurrence (also called crowded scene) and inter-class co-occurrence. Because different object categories usually contain different discriminative features, the problem of inter-class co-occurrence is relatively easier to solve than the one of intra-class co-occurrence. 
In particular, due to the incredible contribution of deep neural network and huge training data \cite{Reference2,Reference3,Reference4,Reference13}, significant improvements have been achieved in object detection \cite{Reference5,Reference8,Reference9,Reference10,Reference14} based on the extremely accurate per-category classification in recent years, thus inter-class co-occurrence is not a big deal anymore for object detection.
However, intra-class co-occurrence still remains as a challenging problem, and most of the popular detectors fail when they come to multiple overlapping objects in crowded scenes.

\begin{center}
\begin{figure}[t]
%\minipage{1.0\textwidth} 
\centering
\includegraphics[height=10.0cm,width=8.0cm]{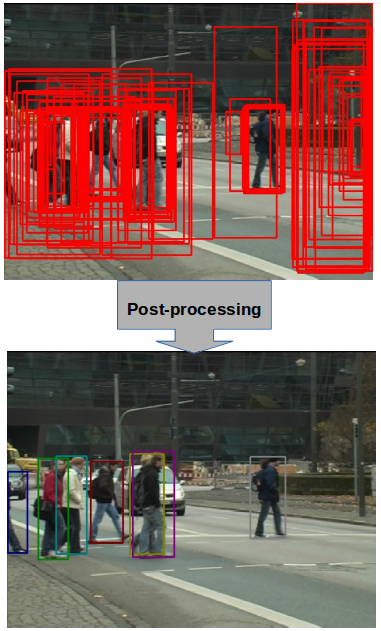}
%\endminipage \hfill
%\vspace{-1em}
\caption{Detection proposals to final bounding boxes: for object detection, on the top, (100) detection proposals (generated by Faster-RCNN) detector, will pass through a post-processing step (GreedyNMS is used by default), and acquire the final bounding boxes as shown at the bottom.}
\label{proposal_box}
\end{figure}
\end{center}

%% 2.2 traditional methods doesn't work->greedNMS
In order to understand the essence of the problem brought by intra-class co-concurrence in multi-object detection, it is necessary to know the work principle and evaluation criteria of object detection. With the developments for object detection in many years, in general, a common pipeline for object detection includes three stages: proposals generation, object detection, and post-processing (using GreedyNMS by default)~\cite{Reference21,Reference5,Reference8,Reference9}. Fig~\ref{proposal_box} shows an example from the detection proposal to the final bounding boxes.  Meanwhile, for evaluation, Mean Average Precision (MAP) is used as the standard criterion~\cite{Reference12,Reference19}. The reason that detectors fail in crowded scenes is that several objects are often closed by each other and heavily occluded. The post-processing step with GreedyNMS is not suitable in such scenarios as it is only capable of achieving good performance for isolated objects in sparse scenes. Specifically, the de facto used GreedyNMS usually suppresses harshly, leading to many missed detections in crowded scenes. Despite this limitation, GreedyNMS is efficient and achieves good performance on sparse scenes, so it has long been the default post-processing step in object detection. 

%% 2.3 in order to overcome the drawback of traditional method. key problem need to be solve

%% 2.4 briefly explain our proposed method

%% 2.5 why it works and can solve the problem.
Targeting a more robust detector for multi-object detection, and overcoming the drawback of GreedyNMS, we propose a new NMS algorithm called Pairwise-NMS. Besides efficiently handling isolated objects like GreedyNMS does, it also takes the heavily occluded case into consideration, and dramatically improves the detection performance in crowded scenes. Specifically, for dealing with two nearby proposals, Pairwise-NMS inherits the GreedyNMS framework when the intersection over union (IoU) is smaller than a pre-defined NMS threshold $N_t$, since GreedyNMS is efficient and accurate in this case. However, when the IoU is larger than the NMS threshold $N_t$, it uses a pairwise-relationship network to predict how many objects the two proposals contain, which helps to handle the multiple overlapping objects effectively. As can be seen in the workflow of Pairwise-NMS in Figure~\ref{fig1:workflow}, comparing to GreedyNMS, Pairwise-NMS is essentially a more general form to handle multi-object detection. To sum up, the contributions of this paper are as follows.

\begin{itemize}
    \item An end-to-end Pairwise-NMS relationship network is proposed for replacing the GreedyNMS, which achieves better performance for object detection without losing efficiency.
    
    \item We evaluate the proposed method on three public datasets, both qualitative and quantitative results demonstrate that our method performs better than GreedyNMS especially in the crowded scene. The proposed method also achieves better performance than the recent Soft-NMS~\cite{Reference26}. %on PETS dataset.
    
    \item Thanks to its flexibility, Pairwise-NMS can be integrated into learning based detectors (e.g., Faster-RCNN and DPM) neatly, and pave the way for the end-to-end learning detectors.
    \end{itemize}
The rest of this paper is organized as follows. Section~\ref{relatedworks} reviews the related work of post-processing for object detection. Section~\ref{methodology} introduces the methodology and the details of the proposed method. Section~\ref{sec:experiments} presents comprehensive experimental results on three public datasets as well as the ablation study. Section~\ref{sec5:conclusion} concludes the paper with future research interests.

%------------------------------------------------------------------------- 

\section{Related works}
\label{relatedworks}
\subsection{Non-Maximum Suppression (NMS)}

\textbf{Sub-Optimal Solver.} 
From the mathematical perspective, non-maximum suppression can be viewed as a quadratic problem. In object detection, the inverse scores of the proposals are treated as the unary terms and IoUs between the proposals are treated as the binary terms~\cite{Reference41}. The scores are generated by detector based on the encoding of localizations between proposals and ground-truth bounding boxes. The role of GreedyNMS is to assign zero if the IoU between two proposals is smaller than a NMS threshold or infinite vice versa. By minimizing an energy function, the problem can be solved through a quadratic optimization solver. As pointed out by previous works~\cite{Reference20,Reference22,Reference23,Reference24}, instead of using GreedyNMS, which greedily iterates and acquires local minimum, using other solvers can provide better local minima. While theoretically appealing, there is an increasing in the complexity in both memory and time by using quadratic optimization solvers, especially when the number of proposals is large \cite{Reference20,Reference22,Reference23,Reference24}. %with the slight gains in performance that are achieved. 
%As the number of proposals increases the cost rises dramatically and to the point that it is rarely practical to anything other than GreedyNMS.

\vspace{2mm}
\noindent \textbf{Message Passing.}
Targeting at object detection in crowded scenes,
Rasmus \textit{et al.}~\cite{Reference25} propose to use a latent structured SVM to learn the weights of affinity propagation clustering between proposals. By passing messages between windows, the merging step can be executed by using clustering methods, and proposals can be re-weighted through minimizing energy function. This kind of method can handle cases which contain two objects that are very close to each other. However, it is hard to distinguish the boundary between one object and two objects. 

%------------------------------------------------------------------------- 
\vspace{2mm}
\noindent \textbf{Soft-NMS.}
Both sub-optimal solver and message passing methods belong to the branch of improving the binary term, \textit{i.e.} re-weight. There are several works focus on improving the unary term, that is re-score, of the proposals. One of the interesting works is Soft-NMS~\cite{Reference26}.  Comparing to GreedyNMS, Soft-NMS decreases the detection scores as an increasing function of IoUs between the target proposals and the overlapped ones, rather than setting scores of the overlapped proposals to zero. Although just one line of code is added, some gains in performance among several detection datasets were  acquired. To be specific, a linear or gaussian function is employed to re-score the non-maximal proposals rather than to totally suppress them. The drawback of this method is that it has to manually tailor a corresponding function to alleviate GreedyNMS. This is troublesome since the balance between recall and precision is  usually trade-off and data driven. Furthermore, as showed in the  experiment section, it can not handle the heavily occluded cases very well.

%------------------------------------------------------------------------- 

\begin{center}
\begin{figure*}[t]
%\minipage{1.0\textwidth} 
\centering
\includegraphics[height=8.0cm]{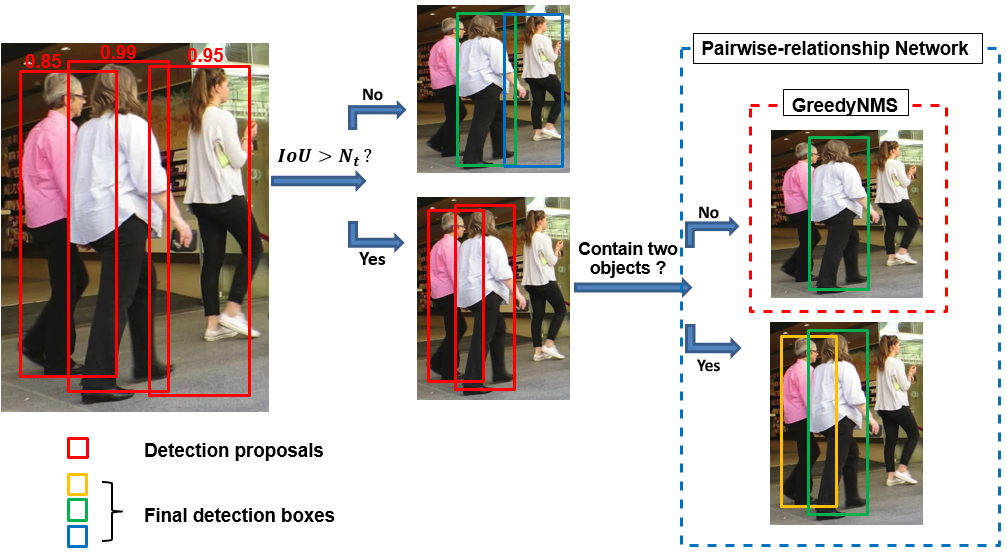}
%\endminipage \hfill
%\vspace{-1em}
\caption{Workflow of Pairwise-NMS. Left: Detection proposals with confident scores are acquired by the detector, the one with highest score with be picked as the final detection result (the middle one, also called `the maximal' proposal), and other proposals which are overlapped with the `the maximal' one will be checked (also called `non-maximal' proposals. Middle: The proposal pairs (between the maximal one and non-maximal ones) will flow into two branches depend on if their IoUs are larger than NMS threshold $N_t$. Right: For the heavily overlapped proposal pairs (in which IoU of the proposal pair is larger than $N_t$), GreedyNMS will supress the non-maximal proposal without consideration. Pairwise-NMS will smartly decide if the non-maximal proposal should be kept depend on the number of objects that the proposal pairs contain. }
\label{fig1:workflow}
\end{figure*}
\end{center}

\vspace{2mm}
\noindent \textbf{ConvNet NMS.}
Very recently, Jan Hoang \textit{et al.}~\cite{Reference27,Reference28} propose to use a neural network to re-score the proposals. Being different from other works, they argue that NMS should be totally replaced with the convolutional neural network. The approach in ~\cite{Reference27,Reference28}  can be used to learn an end-to-end solution for object detection. %, but it does not explicitly learn the pairwise relationship between the two nearby proposals. 
However, we note that for the for sparse scenes, GreedyNMS performs well and efficient. Hence  GreedyNMS should not be totally removed from an object detection pipeline. In addition, Learning-NMS~\cite{Reference28} requires a large memory cost, so it limits the number of the sampling proposals for training the network. 
%------------------------------------------------------------------------- 

\vspace{2mm}
\noindent \textbf{Pairwise Prediction.} 
Based on the same application scenario as~\cite{Reference25}, Tang \textit{et al.}~\cite{Reference29} use a SVM to handle the object pairs in the crowded scene, but with manually designed features. It is hard to handle the case when two nearby objects with the similar appearance.  

Compare to previous works, our work also focuses on predicting object pairs like~\cite{Reference29}, but we employs a deep learning based network. In addition, rather than throwing away NMS framework totally like Learning-NMS~\cite{Reference28},
the good merits of NMS are retained while the hard constraints are mitigated. Specifically, we utilize the features of the region of interest (ROI) to pass messages between pairs.  After the pairwise relationship matrix of the detections within one image was learned, only two lines of codes are added to improve the GeedyNMS (Please refer to Section~\ref{sec:couple_nms} for details). The proposed method is also simple as Soft-NMS~\cite{Reference26}, but it provides a more robust solution for multi-object detection under heavy occlusions.

\section{Methodology}
\label{methodology}

\subsection{Overview}
\label{sec:overview}
%% from problem to the proposal
The standard way to evaluate the inference performance of an object detection method is via mean average precision (MAP). Therefore, both precision and recall are taken into account. Precision is defined as the ratio between the number of true positives and the number of detections, whereas recall is the ratio between the number of true positives and the number of ground-truth bounding boxes. The goal of detection to achieve good performance is to acquire more true positives and less false positives. As the post-processing step of detection, GreedyNMS solely relies on the detection scores to pick the highest-scored proposals while suppressing the overlapping non-maximal proposals in the surrounding. In sparse scenes, GreedyNMS is capable of achieving good performance as there are only very few overlapping objects, in which cases, the highest-scored proposals will be retained to assign to the target objects. However, in crowded scenes, several objects usually are very close to each other. As a result, a bunch of proposals which are likely to be true positives always have similar scores and form a cluster. In this case, GreedyNMS will inevitably lead to a sacrifice between precision and recall, and dramatically harm the overall performance.

In order to generate a more robust detector to handle the general scenes, a new NMS algorithm for post-processing of object detection is required. There are three points should be taken into consideration:

%\vspace{-2mm}
\begin{itemize}
\item  The algorithm is not computationally expensive.

%\vspace{-1mm}
\item  Heavily occluded objects should be taken into considerations.
%\vspace{-1mm}
\item It does not bring in more false positives when improving the recall.
\end{itemize}
%\vspace{-2mm}

In this paper, a new NMS framework is explored, that is Pairwise-NMS. Unlike GreedyNMS, Pairwise-NMS can hold the detection boxes for the nearby objects even when they heavily occluded with each other. Through neatly couping with GreedyNMS without losing efficiency, Pairwise-NMS provides a robust and general solution to the post-processing step for object detection.

%===========================================================

\begin{center}
\begin{figure*}[t]
\centering
\includegraphics[height=6.0cm]{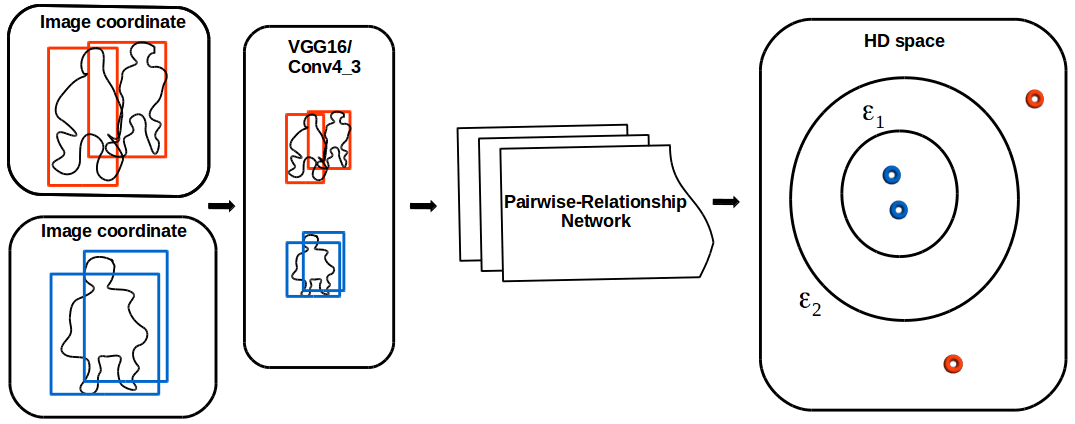}
%\vspace{-1em}
\caption{ Mapping two nearby proposals from image coordinate to high-dimension space. From left to right: The overlapped proposal pairs in image coordinates are first fed into the backbone network (VGG16), their corresponding features extracted from layer Conv4-3 are formed the RoI feature pairs, and used as the inputs of our pairwise-relationship network. Each pair will be mapped into the high-dimention (HD) space through the neural network, and the proposal pair which contain two objects will have a large distance in HD space than the proposal pair which contains zero/one object. }
\label{fig2:pairwise_relationship_network}
\end{figure*}
\end{center}

\subsection{Pairwise-NMS}
Fig~\ref{fig1:workflow} is the workflow of Pairwise-NMS. The algorithm is fed with the detection proposals before NMS as inputs. It is then divided into two branches based on NMS threshold $N_t$. For the top branch, when the IoU of two overlapping proposals is smaller than $N_t$, Pairwise-NMS inherits from GreedyNMS directly because it performs well and very efficient. For the bottom branch, when the IoU of two overlapping proposals is larger than $N_t$, GreedyNMS has a poor recall. Instead of suppressing all of the surrounding non-maximal proposals as GreedyNMS does, the proposal pairs will be fed into a pairwise-relationship network. It will make the Pairwise-NMS to know whether the proposal pairs contain two objects or other cases ( contain zero or one object). If the proposal pair contains two objects, then both of the proposals should be kept no matter how close they are. If the proposal pair contains zero or one object, then the non-maximal proposal would be merged like GreedyNMS does. By doing so, multiple overlapping objects can also be handled effectively. Therefore, Pairwise-NMS can consistently improve the recall and MAP, and considerably increase the performance especially in heavily occluded scenes. 

%------------------------------------------------------------------------- 
%\vspace{-0.5mm}

\subsection{Pairwise Relationship Network}

The goal of pairwise-relationship network is to tell Pairwise-NMS how many objects the two overlapping proposals contain, and let the Pairwise-NMS to decide if the non-maximal proposals should be merged or not. It is challenging to differentiate two heavily overlapping objects which belong to the same category in an image, especially when they have similar shapes as well as appearances. Based on this observation, the pairwise-relationship network is designed. We expect that after mapping, two closed objects in 2D image coordinate will have a large distance in high dimensional space, as shown in Fig~\ref{fig2:pairwise_relationship_network}.
Then, based on the distance that two proposals have in high-dimension (HD) feature space, it can easily infer whether the two proposals contain two objects or not. The input of pairwise-relationship network are features of the proposals from the backbone network (here we use VGG16~\cite{Reference7} as an example, because its simplicity, while it can be replaced with a deeper network structure like ResNet~\cite{Reference13}, GooleNet ~\cite{szegedy2016rethinking} ), and the output is a distance value between the two HD vectors learned from the pairwise-relationship network.
%------------------------------------------------------------------------- 

\subsection{Define Pairs}
\label{sec3.3: define_pairs}
Considering how many objects two proposals contain and whether they are close to each other (Specifically, if IoU of two proposals is larger than NMS threshold, then we define them as ``nearby" or vice versa), there are six cases as shown in Table~\ref{table1:pair_relation}. Case1$\sim$case3 are not under considerations due to the two proposals are not overlapped with each other. For better elucidation, we define case4 and case5 as similar pairs and case6 as dissimilar pairs regarding how many objects the two proposals contain. For similar pairs, in which two proposals contain zero or one object, they should be merged into one bounding box. For dissimilar pairs, where two proposals contain two objects, both of the two proposals should be kept. The pairwise-relationship network mainly focus on predicting if two proposals are similar pairs or dissimilar pairs.

\setlength{\tabcolsep}{4.0 pt}
\begin{table}[t]
\begin{center}
\caption{Relationships between two proposals}
\label{table1:pair_relation}
\scalebox{1.2}{
\begin{tabular}{cccc}
\hline\noalign{\smallskip}
$B_i, B_j$ & nearby or not & objects & constraints in HD-space \\
\noalign{\smallskip}
\hline
\noalign{\smallskip}
case1  & no  & 0 & $\parallel Z_i -Z_j \parallel \geq \epsilon_2$ \\
\noalign{\smallskip}
case2  & no  & 1 & $\parallel Z_i -Z_j \parallel \geq \epsilon_2$  \\

\noalign{\smallskip}
case3  & no  & 2 & $\parallel Z_i -Z_j \parallel \geq \epsilon_2$  \\

\noalign{\smallskip}
case4  & yes & 0 & $\parallel Z_i -Z_j \parallel \leq \epsilon_1$  \\

\noalign{\smallskip}
case5  & yes & 1 & $\parallel Z_i -Z_j \parallel \leq \epsilon_1$  \\

\noalign{\smallskip}
case6  & yes & 2 & $\parallel Z_i -Z_j \parallel \geq \epsilon_2$  \\
\hline
\end{tabular}
}
%\vspace{-1em}
\end{center}
\end{table}
\setlength{\tabcolsep}{1.4pt}
%------------------------------------------------------------------------- 

\begin{center}
\begin{figure*}[t]
\centering
\includegraphics[height=6.0cm]{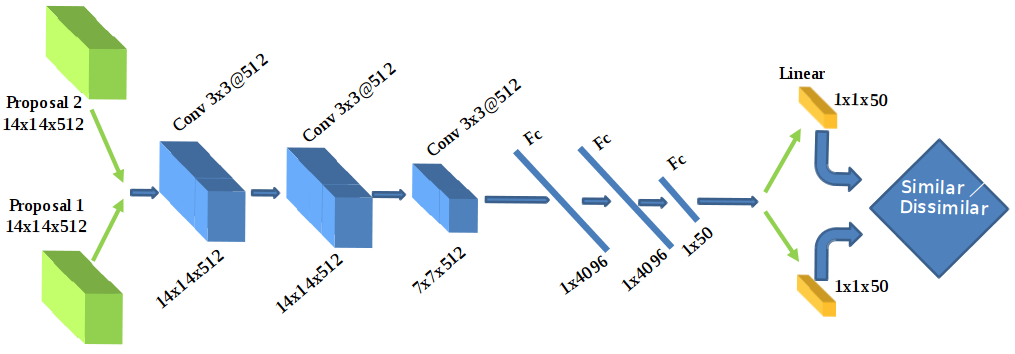}
\caption{Pairwise-relationship network structure. From left to right: The 512-dimension RoI feature pairs are used as input, after three Convolution layers, three FC layers, and one Linear layer (global average pooling\cite{Reference37} is used in our experiments), they are mapped into a 50-dimension space. Relu and Batch Normalization are applied all of the layers except the last one.}
\label{fig3:network_structures}
\end{figure*}
\end{center}

%\vspace{-1mm}
\subsection{Network and Loss}

\noindent
\textbf{Features of Proposal Pairs} The goal of pairwise-relationship network is to learn the relationship between two overlapping proposals. The inputs to the network are the corresponding ROI features of the proposal pairs. We extract ROI features from the layer Conv4\_3 of VGG16~\cite{Reference7}, the size of ROI feature is 1/8 of the corresponding proposal in raw image. The reason for sampling ROI features from Conv4\_3 is because it is small enough, which is good for the training speed and memory cost, and meanwhile carrying enough feature information that the network requires. For the purpose of training the network using the batch operation, ROIAlign~\cite{Reference11} is used to transfer the ROI proposals to the 14 $\times$ 14 fixed size tensors. 

\vspace{2mm} \noindent \textbf{Network Layers}
As shown in Fig~\ref{fig3:network_structures}, the main body of the network structure is composed of three convolutional layers, three fully connected layers and a global average pooling layer~\cite{Reference37}. Each convolutional layer is followed by a Relu~\cite{Reference3} and Batch Normalization~\cite{Reference40} layer. From Conv2 to Conv3, Max-pooling is used. The feature dimension for convolutional layers is 512, and the kernel size and stride are 3 and 1, respectively. In the last FC layer, the dimension of the feature vector is reduced to 50, which is the dimension of the HD space being used to calculate the distance of the two mapped vectors.

%------------------------------------------------------------------------- 

\vspace{2mm} \noindent \textbf{Global Average Pooling}
Global average
pooling(GAP)~\cite{Reference37} is employed after the last FC layer to transfer the tensor to a HD vector. We also conduct some tests to replace the GAP layer with FC layer. Our experimental results demonstrate that GAP layer has a better performance than FC layer for the task, and a brief analysis is provided in Section \ref{sec4.4:ablation}

\vspace{2mm} \noindent 
\textbf{Loss}
We use L1-norm loss because it converges fast, and the loss function is defined as following:
% \begin{equation}
% \label{formula1}
%    L(x_1,x_2)= y\cdot|f(x_1)-f(x_2)|+(1-y)\cdot\max(0,\lambda -|f(x_1)-f(x_2)|)    
% \end{equation}

\begin{equation} \label{formula1}
\begin{split}
L(x_1,x_2)= & y\cdot|f(x_1)-f(x_2)|+ \\
 & (1-y)\cdot\max(0,\lambda -|f(x_1)-f(x_2)|)  
\end{split}
\end{equation}

In Formula~\ref{formula1}, $x_1$ and $x_2$ are ROI proposals, the label $y=1$ when the two proposals belong to similar pairs (zero or one object) and $y=0$ when the two proposals belong to dissimilar pairs (different objects). $f(x_1)$ and $f(x_2)$ are HD vectors learned from the pairwise-relationship network. A margin $\lambda =1$ is used to push the dissimilar pairs away from the similar pairs. $|f(x_1)-f(x_2)|$ is the $L_1$ distance between the two HD vectors. For similar pairs, the distance value between two HD vectors would be forced to approach to 0, while for dissimilar pairs, the distance value between two HD vectors would be forced to approach to 1.

\vspace{2mm} \noindent \textbf{Training and Inference}
For training, the optimizer is stochastic gradient descent (SGD) with momentum \cite{Reference3}. For all of the experiments, the following setting is adopted, where learning rate, momentum, weight decay and batch size are 1e-4, 0.9, 1e-5 and 1 respectively. 
For inference, on each image, a symmetric and sparse pairwise relationship matrix is acquired for the detection proposals, which record the pairwise-distance of any two detection proposal within the image, 
and which is used for couping with GreeyNMS. ON all of the datasets, the total sample proposal pairs during training on every image are 64 for proposals generate by DPM\cite{Reference21} and 32 for proposals generated by Faster-RCNN\cite{Reference9}, and the ratio of dissimilar pairs and similar pairs is 1:3.

% Training epochs and sampling strategies are related to scales of training sets, 
%-----------------------------------
%-----------------------------
%------------------------------------
\setlength{\tabcolsep}{12pt}
\begin{table*}[t]
\small
\begin{center}
\caption{Experiment setting of Pairwise-NMS and GreedyNMS on MOT15 and TUD-Crossing}
\label{table:expsetting}
\scalebox{1.0}{
\begin{tabular}{l|c|c|c|c|c|c|c|c|c|c}
\hline\noalign{\smallskip}
$E_t$ &  0.5 &0.55 & 0.6 & 0.65 & 0.7 & 0.75 &0.8 & 0.85 & 0.9 & 0.95 \\
\noalign{\smallskip}
\hline
\noalign{\smallskip}
$N_t$  & 0.55 & 0.55 & 0.55 & 0.6 & 0.65 & 0.7 & 0.8 & 0.85 & 1 & 0.95\\
\noalign{\smallskip}
\hline
\noalign{\smallskip}
$D_t$ & 1.25 & 1.25 & 1.25 & 1.4 & 1.3 & 0.65 & 0.9 & 0.5 & 0 & 0.2\\
\hline
\end{tabular}
}
%\vspace{-1em}
\end{center}
\end{table*}
\setlength{\tabcolsep}{1.4pt}

\subsection{Coupling with GreedyNMS}
\label{sec:couple_nms}
Just like Soft-NMS\cite{Reference26}, Pairwise-NMS can also be coupled with GreedyNMS neatly. The following is the coupling code of Pairwise-NMS, after the pairwise relationship maxtrix is learned through the pairwise-relationship network. The colored parts are the differences between GreedyNMS and Pairwise-NMS. As can be seen, in GreedyNMS, all of the non-maximal proposals would be suppress, regardless how many objects it has in the surrounding area, which will inevitably cause the performance drop under the heavily occluded cases. On the contrary, our method provide an effective complementary to the GreedyNMS, that is, even when IoU of the non-maximal proposal with the selected proposal is larger than NMS threshold, those two proposals (the proposal pair) will be passed into the pairwise-relationship network, and let the network automatically decided how many objects the two proposals contain. Compared with the `blind' suppression of GreedyNMS, Pairwise-NMS is much flexible and robust for multi-object detection. 
And comparing to GreedyNMS, just one line of code is added, thus no efficiency lose.

%-------------------------------------

\renewcommand{\algorithmicrequire}{\textbf{Input:}}   %Use Input in the format of Algorithm
\renewcommand{\algorithmicensure}{\textbf{Output:}}  %Use Output in the format of Algorithm

\begin{algorithm}[t]
    \caption{Pusedo code of Pairwise-NMS}
\begin{algorithmic}[2]
%------------------
\REQUIRE
$\beta=\{b_1,...b_N \}$, $S=\{s_1,...s_N \}$, $N_t$ \\ 
$\beta$ is the list of initial detection boxes \\
$S$ contains corresponding detection scores\\
$N_t$ is the NMS threshold

$D_t$ is the distance threshold

\label{coupling_pairwise_nms}
\end{algorithmic}

\Begin
{   $D \leftarrow \{\}$ \\
     \While{$\beta \neq empty $}
     {%
         $m \leftarrow argmax \hspace{0.5em} S$ \\
         $M \leftarrow b_m$ \\
         $D \leftarrow D \cup M  ; B \leftarrow B - M$
    
      \For{$b_i$ in $\beta$}
      {
        \begin{elaboration_red}
        \If {$iou(M,b_i) \geq N_t$}
        {$\beta \leftarrow \beta - b_i ; S \leftarrow S - S_i $}
        \hfill \textbf{GreedyNMS}
        \end{elaboration_red} 
        
        \vspace{0.2cm}
        \begin{elaboration_green}
        \If {$iou(M,b_i) \geq N_t \wedge \\
         pairDist(M,b_i) \leq D_t $
        }
        {$\beta \leftarrow \beta - b_i ; S \leftarrow S - S_i $}
        \hfill \textbf{Pairwise-NMS}
        \end{elaboration_green}
        
      }
    }
}
\end{algorithm}

%------------------------------------------------------------------------

\section{Experiments}
\label{sec:experiments}
\subsection{Experiment Glimpse and Evaluation}
\label{sec:evaluation}

\subsubsection{Experiment Glimpse}
\noindent 
In this part, we provide extensive experiments on three different datasets, including MOT15, TUD-Crossing and PETS. The experimental results firstly demonstrate how our Pairwise-NMS outperforms GreedyNMS in crowded scenes with occlusions to different extents, especially, the gains with the increasing of the occlusions that Pairwise-NMS over GreedyNMS will be emphasized. Moreover, the comparison with Soft-NMS demonstrate our method performs better in very crowded scenes.

% \setlength{\tabcolsep}{4pt}
% \begin{table}[t]
% \begin{center}
% \caption{Experiment setting of Pairwise-NMS and GreedyNMS on MOT15 and TUD-Crossing}
% \label{table:expsetting}
% \scalebox{1.2}{
% \begin{tabular}{l|c|c|c|c|c}
% \hline\noalign{\smallskip}
% $E_t$ &  0.5 &0.55 & 0.6 & 0.65 & 0.7  \\
% \noalign{\smallskip}
% \hline
% \noalign{\smallskip}
% $N_t$  & 0.55 & 0.55 & 0.55 & 0.6 & 0.65 \\
% \noalign{\smallskip}
% \hline
% \noalign{\smallskip}
% $D_t$ & 1.25 & 1.25 & 1.25 & 1.4 & 1.3 \\
% \noalign{\smallskip}
% \noalign{\smallskip}
% \hline\noalign{\smallskip}
% $E_t$ & 0.75 & 0.8 & 0.85 & 0.9 & 0.95 \\
% \noalign{\smallskip}
% \hline
% \noalign{\smallskip}
% $N_t$ & 0.7 & 0.8 & 0.85 & 1 & 0.95\\
% \noalign{\smallskip}
% \hline
% \noalign{\smallskip}
% $D_t$ & 0.65 & 0.9 & 0.5 & 0 & 0.2\\

% % \hline\noalign{\smallskip}
% % $E_t$&  0.75 &0.8 & 0.85 & 0.9 & 0.95 \\
% % \noalign{\smallskip}
% % \hline
% % \noalign{\smallskip}
% % $N_t$  & 0.7 & 0.8 & 0.85 & 1 & 0.95\\

% % $D_t$ & 0.65 & 0.9 & 0.5 & 0 & 0.2 \\
% \hline
% \end{tabular}
% }
% %\vspace{-1em}
% \end{center}
% \end{table}
% \setlength{\tabcolsep}{1.4pt} 

\subsubsection{About Evaluation}
\noindent There are two parameters to be considered during the execution of NMS related to the final detection performance. One of the parameters is NMS threshold $N_t$, and it determines to what extent the surrounding non-maximal proposals should be suppressed by the local maximal proposal. Another parameter is the evaluation threshold $E_t$, and it determines to what extent a detection bounding box could be true positive (if only one detection box is aligned to the object) or false positives (if multiple detection boxes are aligned to the same object).

When a small evaluation threshold (for example $E_t=0.5$) is used, it is much easier for the detection proposals to be true positives, since the only condition to be satisfied is that their IoUs with ground truth boxes are bigger or equal than 0.5. Meanwhile, due to the constraint that the evaluation script asking for at most one proposal to align the same object, as a result, it is prone to high recall and lower precision. On the contrary, when a large evaluation threshold (for example $E_t=0.95$) is used, it is more difficult for the detection proposals to be true positives because it relies on the detector to perform an extremely accurate localization. As a result, it will lead to high precision and low recall.

In order to perform an objective and comprehensive evaluation, especially in multi-object detection which often contains heavy occluded cases, we use the same evaluation criterion as Microsoft COCO evaluation~\cite{Reference12}, and with different evaluation thresholds to reflect the strength of Pairwise-NMS for handling heavy occlusions in crowded scenes.
%-------------------------------------------------------------------------

\subsection{Results on MOT15}
For each evaluation criterion, one optimal NMS threshold $N_t$ and one distance threshold $D_t$ are chosen for GreedyNMS and Pairwise-NMS respectively. The specific settings for MOT15 and TUD-Crossing are listed as Table~\ref{table:expsetting}. And the setting for PETS are following \cite{Reference28} and will be explained in the Section~\ref{sect:pets}

MOT15 benchmark~\cite{Reference38} is a public dataset, which collects 11 sequences for multi-object detection and tracking from different cities. It is challenging due to the heavy occlusions, dramatic illumination changes and clustering background. MOT15 training set is used as it provides the ground-truth annotations, which facilitates the verification of the proposed Pairwise-NMS framework. The whole training set contains 5500 images, and is split with a ratio 5:1:4 for training, validation and test. For object detector, we use the Faster-RCNN code implemented by Ross Girshick\cite{Reference9}. That detector was originally trained on COCO~\cite{Reference12} and PASCAL VOC~\cite{Reference19}, both of which lack significant numbers of occluded objects. We therefore fine-tune the detector on the MOT15 training set as mentioned above. For all of the experiments, pairwise-relationship network was trained on training set and parameters were fine-tuned on validation set only.

\setlength{\tabcolsep}{4.0pt}
\begin{table}[t]
\small
\begin{center}
\caption{AP of optimal GreedyNMS vs Pairwise-NMS on MOT15}
\label{table2:mot15_ap}
\scalebox{1.0}{
\begin{tabular}{llll}
\hline\noalign{\smallskip}
Evaluation threshold & GreedyNMS & Pairwise-NMS & Outperform \\
\noalign{\smallskip}
\hline
\noalign{\smallskip}
AP @ 0.5  & 74.11 & \textbf{74.15} & 0.04 \\

\noalign{\smallskip}
AP @ 0.55 & 76.15 & \textbf{76.21} & 0.06 \\

\noalign{\smallskip}
AP @ 0.6  & 65.66 & \textbf{65.82} & 0.16 \\

\noalign{\smallskip}
AP @ 0.65 & 57.86 & 57.86 & 0.0  \\

\noalign{\smallskip}
AP @ 0.7  & 45.68 & \textbf{46.41} & 0.73 \\

\noalign{\smallskip}
AP @ 0.75 & 29.45 & \textbf{30.41} &0.96\\

\noalign{\smallskip}
AP @ 0.8  & 16.70 & \textbf{16.82} & 0.12 \\

\noalign{\smallskip}
AP @ 0.85 & 5.17  & \textbf{5.26}  & 0.09  \\

\noalign{\smallskip}
AP @ 0.9  & 0.87  & 0.87  & 0.0 \\

\noalign{\smallskip}
AP @ 0.95 & 0.02  & 0.02  & 0.0 \\
\hline
\end{tabular}
}
%\vspace{-1em}
\end{center}
\end{table}
\setlength{\tabcolsep}{1.4pt}

Table~\ref{table2:mot15_ap} shows the overall performance of GreedyNMS and Pairwise-NMS on test set. We use the same evaluation as Microsoft COCO~\cite{Reference12}, which is $AP_{0.5}$ to $AP_{0.95}$ , as the evaluation criterion. For fair comparison, the optimal NMS threshold was acquired under each evaluation threshold, and based on that, the best AP for GreedyNMS is set down as the baseline. Using the same setup with GreedyNMS, our pairwise-relationship network was trained. The pairwise matrix that is obtained using pairwise-relationship network from inference is then be coupled into GreedyNMS framework as described in Section~\ref{sec:couple_nms}.
As can be seen, Pairwise-NMS outperforms GreedyNMS is eligible for $AP_{0.5}$. However, the gap is enlarged with the increasing of evaluation threshold. For example, when $AP_{0.75}$ is used, Pairwise-NMS beats GreedyNMS about 1 percent for overall performance. This improvement also double proves that our method is good at handling heavy occlusions. 

Although we have achieved consistent improvements compared to GreedyNMS using different evaluation thresholds, the gains of the overall performances seem not large. In fact, it does not like so. After a thorough investigation of datasets which have heavy occlusions including TUD-Crossing and MOT15, we discover that the number of heavily occluded objects is relatively smaller comparing to the total number of ground truth bounding boxes. Besides, the brief analysis about the factors that affect the final evaluation provided in Section~\ref{sec:evaluation} can also provide some hints.

%%==========================================
\begin{center}
\begin{figure}[t]
\centering
\includegraphics[height=7.0cm]{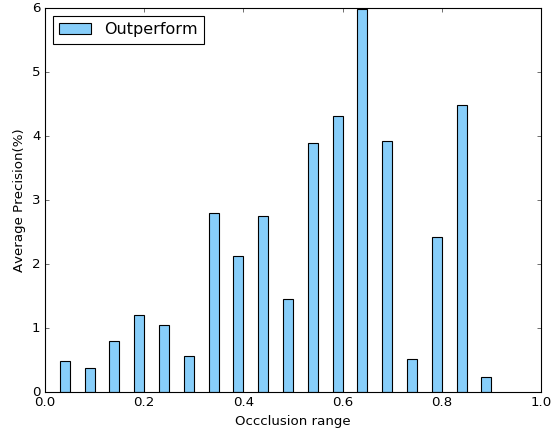}
%\vspace{-1em}
\caption{Outperform percentages of Pairwise-NMS than optimal GreedyNMS under different occlusion ranges on MOT15. Using AP@0.5 as evalution cretion, Pairwise-NMS achive 6\% higher percent than that of GreedyNMS regarding to detection accuracy, and with the increasing of the occlusion range, the gains that Pairwise-NMS outperformed GreedyNMS is enlarged.}
\label{fig4:mot15_occlusion}
\end{figure}
\end{center}
% %%---------------- try3 ---------------
% % Table float box with bottom caption, box width adjusted to content
% \begin{figure}[t]
% \begin{floatrow}
% \ffigbox{%
% \includegraphics[height=5.5cm,width=6cm]{images/fig5_bar_chart1.png}

% % fig5_bar_chart is absolute value
% \vspace{-1em}

% }
% {\caption{Outperform percentages of Pair wise-NMS than optimal GreedyNMS under different occlusion ranges on MOT15
% \label{fig4:mot15_occlusion}}%
% }

% %%----------------- table ------------------ %%
% \capbtabbox{%
% \scalebox{1.0}{
% \begin{tabular*}{6cm}{cccc} \hline 
% \hline\noalign{\smallskip}
% Gt& Greedy& Pairwise& Out- \\
% overlap & NMS & -NMS & perform \\
% \noalign{\smallskip}
% \hline
% \noalign{\smallskip}
% 0.4  - 0.45  & 38.73 &\textbf{40.25} & 1.52\\

% 0.45  - 0.5  & 39.17 &\textbf{39.18} & 0.01\\

% 0.5  - 0.55  & 43.93 &\textbf{44.72} & 0.79\\

% 0.55 - 0.6 & 32.18  & \textbf{33.28} & 1.1 \\

% 0.6  - 0.65  & 29.03  & \textbf{30.0} & 0.97 \\
% 0.65 - 0.7   & 32.17  & \textbf{35.90} & 3.73 \\

% 0.7  - 0.75  & 39.46  & \textbf{39.73}  & 0.27 \\

% 0.75 - 0.8   & 36.78  & 36.78 & 0.0 \\

% 0.8  - 0.85  & 33.65  & \textbf{34.29}  & 0.64 \\

% 0.85  - 0.9  & 6.25 & \textbf{6.90}  & 0.65 \\

% \hline
% \end{tabular*}
% }
% %\vspace{-1em}
% }{%
%  \caption{F1-score of optimal GreedyNMS vs Pairwise-NMS under different occlusion ranges on MOT15}
% \label{table3:mot15_f1}%
% }
% \end{floatrow}
% \end{figure}
% %%====================================

\setlength{\tabcolsep}{4pt}
\begin{table}[t]
\small
\begin{center}
\caption{F1-score of optimal GreedyNMS vs Pairwise-NMS under different occlusion ranges on MOT15}
\label{table3:mot15_f1}
\scalebox{1.0}{
\begin{tabular}{llll}
\hline\noalign{\smallskip}
Gt overlap& GreedyNMS & Pairwise-NMS& Outperform \\
\noalign{\smallskip}
\hline
\noalign{\smallskip}
0.4  - 0.45  & 38.73 &\textbf{40.25} & 1.52\\

\noalign{\smallskip}
0.45  - 0.5  & 39.17 &\textbf{39.18} & 0.01\\

\noalign{\smallskip}
0.5  - 0.55  & 43.93 &\textbf{44.72} & 0.79\\

\noalign{\smallskip}
0.55 - 0.6 & 32.18  & \textbf{33.28} & 1.1 \\

\noalign{\smallskip}
0.6  - 0.65  & 29.03  & \textbf{30.0} & 0.97 \\

\noalign{\smallskip}
0.65 - 0.7   & 32.17  & \textbf{35.90} & 3.73 \\

\noalign{\smallskip}
0.7  - 0.75  & 39.46  & \textbf{39.73}  & 0.27 \\

\noalign{\smallskip}
0.75 - 0.8   & 36.78  & 36.78 & 0.0 \\

\noalign{\smallskip}
0.8  - 0.85  & 33.65  & \textbf{34.29}  & 0.64 \\

\noalign{\smallskip}
0.85  - 0.9  & 6.25 & \textbf{6.90}  & 0.65 \\
\hline
\end{tabular}
}
%\vspace{-1em}
\end{center}
\end{table}
\setlength{\tabcolsep}{1.4pt}
%%-----------------------------------------

For better analysis about the performance of our algorithm, like the previous related papers \cite{Reference26,Reference27,Reference28}, $AP_{0.5}$ for different occlusions are acquired. Here, we use the overlap between ground-truth bounding boxes as the occlusion reference.
Fig~\ref{fig4:mot15_occlusion} shows a full comparison of outperform percentages of Pairwise-NMS than GreedyNMS under different occlusion ranges. As can be seen in Fig~\ref{fig4:mot15_occlusion}, in most cases, Pairwise-NMS beat optimial GreedyNMS. The overall trend is that the heavier the occlusions, the more gains Pairwise-NMS achieve, and the max gains is 6 percent compared with GreedyNMS.

To gain further understanding on the performance of our method under heavy occlusions, F1 scores under different occlusion ranges from 0.4 to 0.9 with interval 0.05 are outlined in Table~\ref{table3:mot15_f1}. As can ben seen, Pairwise-NMS has better F1-score than GreedyNMS in all of the thresholds. Pairwise-NMS outperforms GreedyNMS more than 3.7 percent under occlusion range between 0.65 and 0.7.

%-------------------------------------------------------------------------
\subsection{TUD-Crossing for Generalization}

\begin{center}
\begin{figure}[t]
\centering
\includegraphics[height=6.5cm]{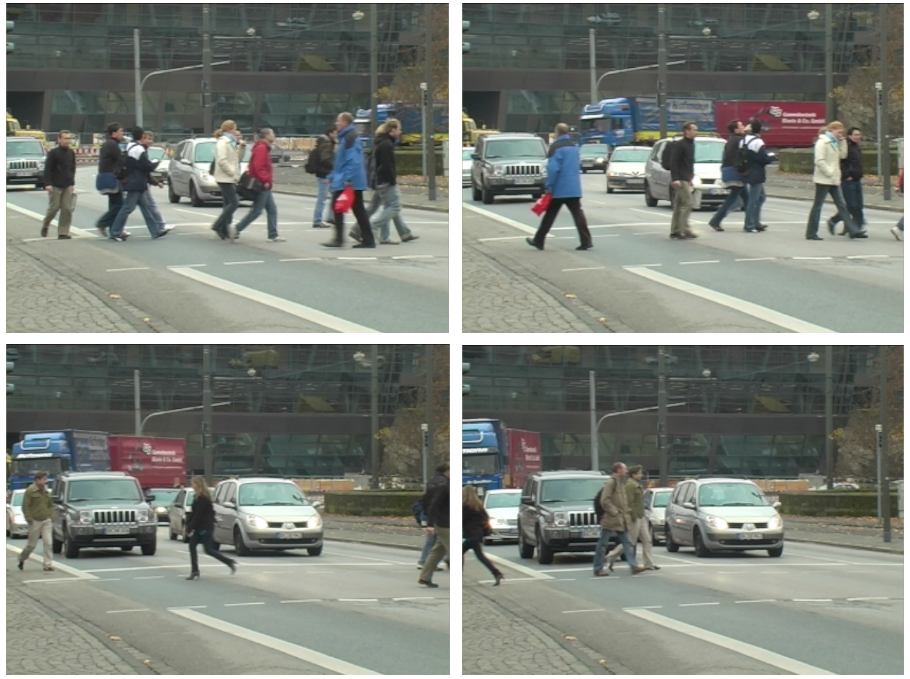}
%\vspace{-1em}
\caption{TUD-Crossing: the 50th, 100th, 150th, 200th frame of  dataset, as can be seen, it is quite challenging as for object detection.}
\label{tud-crossing}
\end{figure}
\end{center}

In order to verify the generalization ability of the proposed method, we use TUD-Crossing~\cite{Reference39} dataset, which is composed of 201 heavily occluded images, to perform the inference based on the trained model which is obtained from MOT15, without any fine-tuning on TUD-Crossing.

In Figure~\ref{tud-crossing}, there are four images with frame no 50, 100, 150, 200 are shown out. Since TUD-Crossing dataset are capture from the side view when pedestrians are coming across the traffic light, as can be seen, in all of the images, there are always heavily occluded cases happen, regardless how many people within the camera frustum. And for the first half of the video sequence, the camera frustum usually contain a relatively higher density. Moreover, the crowding, small scale, fractional and low resolution also make the task more challenging.

The results are shown in Table~\ref{table4:tud_ap}, average precision (AP) is used as the measurement metric. And for most of the evaluation criterions, Pairwise-NMS performs better than GreedyNMS regarding to detection accuracy. Specifically, the heavier occlusions are, the more gains Pairwise-NMS achieves when compared with that of using GreedyNMS, and the maximumal overall gain is 0.95 percent, and considering the ratio of heavily occluded cases just take up small part of the overall bounding boxes, this is a large improvements for object detection. To sum up, the detection results on TUD-Crossing dataset present the similar trend as the one on MOT15, which confirms that our pairwise-relationship network has acquired the ability for handling with the heavy occlusions regarding the texture and pattern changing of the image sequence.

\setlength{\tabcolsep}{4pt}
\begin{table}[t]
\small
\begin{center}
\caption{AP of optimal GreedyNMS vs Pairwise-NMS on TUD-Crossing}
\label{table4:tud_ap}
\scalebox{1.0}{
\begin{tabular}{llll}
\hline\noalign{\smallskip}
Evaluation threshold & GreedyNMS & Pairwise-NMS & Outperform \\
\noalign{\smallskip}
\hline
\noalign{\smallskip}
AP @ 0.5  & \textbf{79.07} & 78.86 & -0.21 \\

\noalign{\smallskip}
AP @ 0.55 & \textbf{75.58} & 75.25 & -0.33 \\

\noalign{\smallskip}
AP @ 0.6  & 69.46 & \textbf{69.47} & 0.01 \\

\noalign{\smallskip}
AP @ 0.65 & 57.95 & \textbf{58.07} & 0.12  \\

\noalign{\smallskip}
AP @ 0.7  & 42.33 & \textbf{42.35} & 0.014 \\

\noalign{\smallskip}
AP @ 0.75 & 24.41 & \textbf{25.36} & 0.95\\

\noalign{\smallskip}
AP @ 0.8  & 12.40 & \textbf{12.56} & 0.16 \\

\noalign{\smallskip}
AP @ 0.85 & 4.02  & \textbf{4.10}  & 0.07  \\

\noalign{\smallskip}
AP @ 0.9  & 0.51  & 0.51  & 0.0 \\

\noalign{\smallskip}
AP @ 0.95 & 0.01  & 0.01  & 0.0 \\
\hline
\end{tabular}
}
%\vspace{-1em}
\end{center}
\end{table}
\setlength{\tabcolsep}{1.4pt}

%-------------------------------------------------------------------------
\subsection{Results on PETS }
\label{sect:pets}
PETS~\cite{Reference42} dataset is another public dataset which consists of eight sub-sequences that are captured from different angles and views. We use the same setup to split the dataset as \cite{Reference28,Reference29}, and the detection proposals before NMS were generated by DPM~\cite{Reference21}. Table~\ref{table6:PETS} provides the statistics of PET dataset, including the sub-sequence splits, number of frames, detection proposals before NMS and the ground truth bounding boxes for training, validation and testing.

A same workflow as MOT15 for training and inference is shared. 
Table~\ref{table5:pets_ap} are the results we obtain using Pairwise-NMS and GreedyNMS with different NMS thresholds, in order to be consistent with~\cite{Reference29}, all of the results, with $AP_{0.5}$ as the evaluation criterion. As can be seen, our method outperforms all of the GreedyNMS results, and beats GreedyNMS with optimal NMS threshold ($N_t=0.4$) more than 1.5 percent. It proves that, as the post-processing step, Pairwise-NMS is a good replacement for GreedyNMS and it can be integrated into any detector.

\begin{center}
\begin{figure}[t]
\centering
\includegraphics[height=6.5cm]{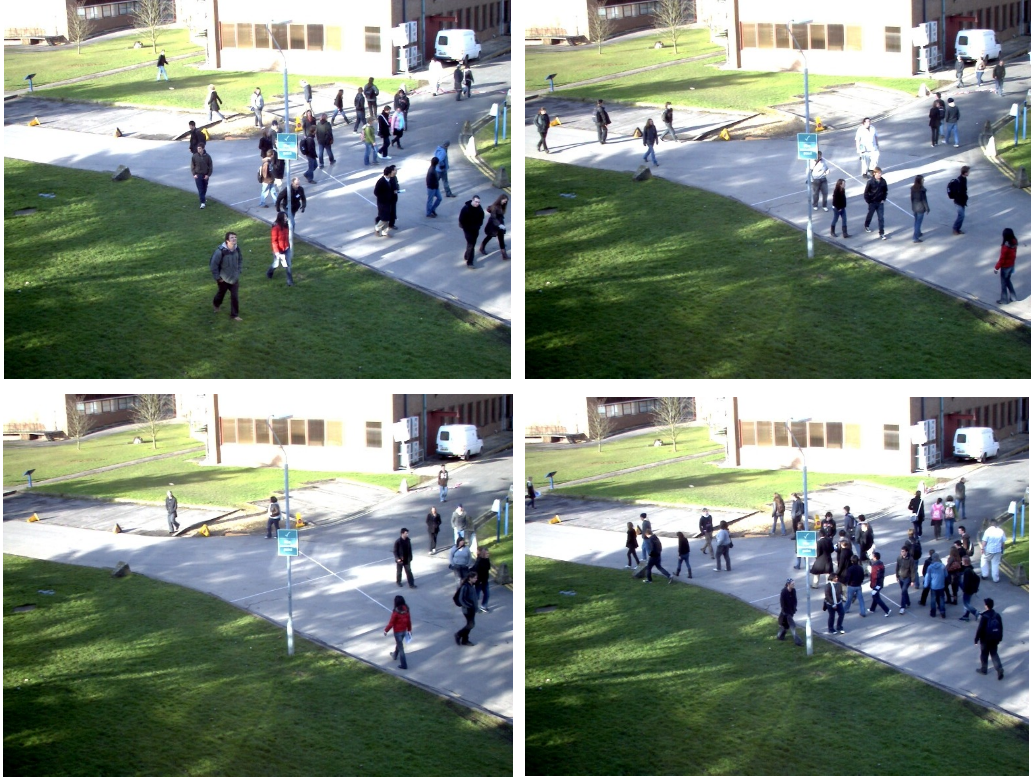}

\caption{PETS test images: The 100th, 200th, 300th, 400th frame of PETS test set (S2L2 sequence) are shown out. As can be seen, the people in this scene is very crowded, quite small and affected by the light conditions, thus is extremely challenging for object detection.}
\label{pets_test}
\end{figure}
\end{center}

\setlength{\tabcolsep}{4pt}
\begin{table}[t]
\begin{center}
\caption{Statistic of PETS Dataset}
\label{table6:PETS}
\scalebox{1.0}{
\begin{tabular}{ccccc}
\hline\noalign{\smallskip}
splits & sequences & frames & proposals & gt boxes \\
\noalign{\smallskip}
\hline
\noalign{\smallskip}
\multirow{2}*{training} & S1L1-1, S1L1-2, S1L2-1  &\multirow{2}*{1696} & \multirow{2}*{5421758} & \multirow{2}*{23107} \\
                        & S1L2-2, S2L1, S3MF1    &                    &                        &    \\   
%\hline
\noalign{\smallskip}
validation & S2L3 & 240 & 727964 & 4376 \\

%\hline
\noalign{\smallskip}

test  & S2L2 &436 & 2112655  &  10292\\
\hline
\end{tabular}
}
\end{center}
\end{table}
\setlength{\tabcolsep}{1.4pt}

%% ----------try -------------------
\setlength{\tabcolsep}{4pt}
\begin{table}[t]
\begin{center}
\caption{AP$_{0.5}$ of Pairwise-NMS vs GreedyNMS on PETS}
\label{table5:pets_ap}
\scalebox{1.0}{
\begin{tabular}{c|c|cccc}
\hline\noalign{\smallskip}
Method &  Pairwise-NMS &  & GreedyNMS & & \\
\hline
\multirow{4}{*}{AP @ 0.5} &\multirow{4}{*}{\textbf{78.1}}  &$N_t>$0.0  & $N_t>$0.1 &$N_t>$0.2 &$N_t>$0.3  \\
%\noalign{\smallskip}

%\noalign{\smallskip}
  & &55.0  &66.0 & 71.4  & 75.0 \\ \cline{3-6} 
  
%\noalign{\smallskip}
& & $N_t>$0.4 &$N_t>$0.5 &$N_t>$0.6 &\\ 

%\noalign{\smallskip}
&  & 76.6& 73.4 &64.8 & \\
  
\hline
\end{tabular}
}
%\vspace{-1em}
\end{center}
\end{table}

\setlength{\tabcolsep}{1.4pt}

%-------------------------------------------------------------------------
\subsection{Comparison with Soft-NMS}
In this section, we conduct the comparison experiments between the proposed Pairwise-NMS and the state-of-the-art method Soft-NMS\cite{Reference26} on PETS test dataset.

Since the most outstanding advantage of Pairwise-NMS is capable of handling with multi-object detection especially under the heavy occluded cases. PETS dataset is chosen for the reason that it is one of the most crowded and widely used datasets for multi-object detection. Thus it should be representative and  is reasonable to reflect the robustness of algorithms.

According to the Soft-NMS~\cite{Reference26} paper, instead of suppressing all of the proposals which have heavy IoUs with the targeting (maximal scored) proposal, Soft-NMS replace the hard NMS threshold used in GreedyNMS with a linear or non-linear function to re-score the overlapped proposals. And the general principle that guiding the re-scoring process is that, the IoU between the two overlapped proposals should be act as a weight for the `score-decay` of the non-maximal one. Either the linear function or Gaussian function could be utilized to execute the re-scoring process depends on their performance regarding to the dataset.

Since for both of the linear function based Soft-NMS and Gaussian function based Soft-NMS, the scores of the non-maximal proposals are continuous penalized, as a result, there will be no proposal removed from each suppression round, which will cause a long excution time compared with GreedyNMS, thus there is a parameter $\theta$ used in practical. The role of $\theta$ is to filter out the proposals which have smaller scores regardless what their originals scores are. By this way, Soft-NMS can keep the same level of efficiency with GreedyNMS. For Gaussian function based Soft-NMS, there is an extra parameter $\sigma$ which is the standard deviation of the Gaussian function, and can be viewed as a `controller` of the distribution shape of the penalty function.

\setlength{\tabcolsep}{10pt}
\begin{table*}[t]
\small
\begin{center}
\caption{Comparison results with Soft-NMS on PETS}
\label{table_soft-nms}
\scalebox{1.0}{
\begin{tabular}{ccccccccc}
\hline\noalign{\smallskip}
methods & thre & tp & fp & dt & gt & rec & prec & AP@0.5\\
\noalign{\smallskip}
\hline
\noalign{\smallskip}
Soft-NMS\_L & 0.0 & 10119 & 606298 & 616417 & 10292 & 98.32 & 1.64 & 37.60\\
\noalign{\smallskip}
Soft-NMS\_L & 0.1 & 8402 & 11992 & 20394 & 10292 & 81.64 & 41.20 & 60.75\\
\noalign{\smallskip}
Soft-NMS\_L & 0.2 & 7601 & 2942 & 10543 & 10292 & 73.85 & 72.10 & \textbf{63.27}\\
\noalign{\smallskip}
Soft-NMS\_L & 0.3 & 6645 & 540 & 7185 & 10292 & 64.56 & 92.48 & 61.94\\

\noalign{\smallskip}
Soft-NMS\_L & 0.4 & 5494 & 192 & 5686 & 10292 & 53.38 & 96.62 & 52.40\\

\noalign{\smallskip}
\hline
\noalign{\smallskip}
Soft-NMS\_G & 0.1 & 7216 & 1383 & 8599 & 10292 & 70.11 & 83.92 & 68.05\\

\noalign{\smallskip}
Soft-NMS\_G & 0.2 & 7543 & 2100 & 9643 & 10292 & 73.29 & 78.22 &    \textbf{70.24}\\

\noalign{\smallskip}
Soft-NMS\_G & 0.3 & 7756 & 3115 & 10871 & 10292 & 75.36 & 71.35 & 69.00\\

\noalign{\smallskip}
Soft-NMS\_G & 0.4 & 7914 & 4561 & 12475 & 10292 & 76.89 & 63.44 & 64.73\\

\noalign{\smallskip}
Soft-NMS\_G & 0.5 & 8012 & 6014 & 14026 & 10292 & 77.85 & 57.12 & 59.84\\

\noalign{\smallskip}
\hline
\noalign{\smallskip}
Pairwise-NMS & 0.7 & 9477 & 66718 & 76195 & 10292 & 92.08 & 12.44 & \textbf{78.11}\\
\hline
\end{tabular}
}
%\vspace{-1em}
\end{center}
\end{table*}

\setlength{\tabcolsep}{1.4pt}

In Table~\ref{table_soft-nms}, on PETS dataset, we list the detection results of Soft-NMS under different settings and that of Pairwise-NMS. Specifically, with Soft-NMS\_L denotes linear function based Soft-NMS, and with Soft-NMS\_G denotes Gaussian function based Soft-NMS. For easy comparison, we unified use the 'thre' in the second column of Table~\ref{table_soft-nms} to denote the parameter $\theta,\sigma$ and $N_d$ employed in linear function based Soft-NMS, Gaussian function based Soft-NMS and Pairwise-NMS, respectively. In addition, as point out before, for Gaussian function based Soft-NMS, both the parameters $\theta$ and $\sigma$ are used, after fine tuning the $\theta$ on linear function based Soft-NMS, we fix it as 0.2, and just take $\sigma$ as the variable. In order to have a comprehensive comparison and understanding about different algorithms under various settings, some key measurements are listed, they are true positives, false positives, number of final detections, number of ground-truth bounding boxes, recall, precision and average precision with evaluation threshold 0.5 (tp, fp, dt, gt, rec, prec, AP@0.5) on the first row.

%%\begin{center}
\begin{figure}[t]
\centering
\includegraphics[height=7.0cm]{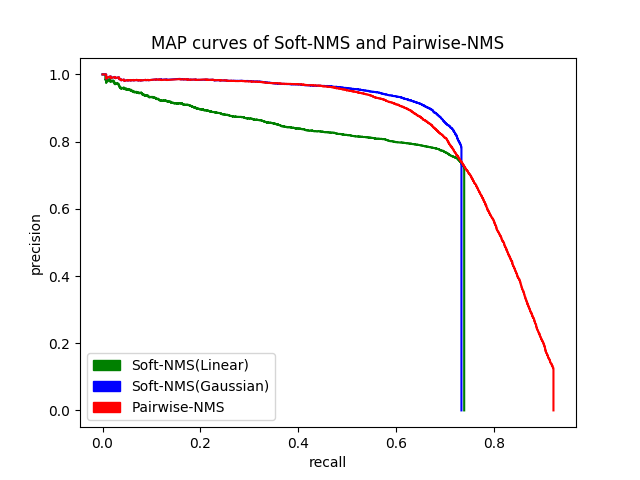}
%\vspace{-1em}
\caption{MAP curves of Soft-NMS and Pairwise-NMS. As can be been, both of the liner function based Soft-NMS and Gaussian function based Soft-NMS have lower performance than that of Pairwise-NMS on PETS dataset regarding to detection accuracy. Form surface, it is struggle for both type of Soft-NMS to reach a good balance between reall and precision on PETS dataset. From essential, we suspect that Pairwise-NMS has superior ablity than Soft-NMS to handle the very crowded cases.}
\label{fig4_nms_curves}
\end{figure}
%%\end{center}

From observing the AP performance of different algorithms, it can be concluded that, both the linear function based Soft-NMS and the Gaussian function based Soft-NMS have lower performance than our method on PETS dataset. Specifically, Soft-NMS\_L achieves highest AP of 63.27 percent, and Soft-NMS\_G achieves highest AP of 70.24. Both of them are far behind from the performance of our proposed Pairwise-NMS. And from watching the recall and precision, it is much clear that, for Soft-NMS, it almost fails in all settings to reach a good trade-off between the precision and recall, which can be viewed as the surface explanation of the unsatisfied APs.
And from the essential, we suspect there are two reasons:
firstly, from the perspective of dataset,
both of the VOC2007 and COCO dataset used by Soft-NMS is relative sparse and contains very few crowed cases, on the contrary, PETS is very crowded and is much challenging. For example, a lot of frames contain above 30 pedestrians, and they are quite small in resolutions and heavily occluded with each other. Secondly, from the perspective of detector, in order to keep the same settings with \cite{Reference27,Reference28,Reference29}, detection proposals generated by DPM (before NMS) are used here. The detection proposals for each image easily reach several thousands. Which may cause more difficulties for Soft-NMS compared to that at most 300 detection proposals are used in the original paper of Soft-NMS\cite{Reference26}

\begin{center}
\begin{figure*}[t]
\centering
\includegraphics[height=9.0cm]{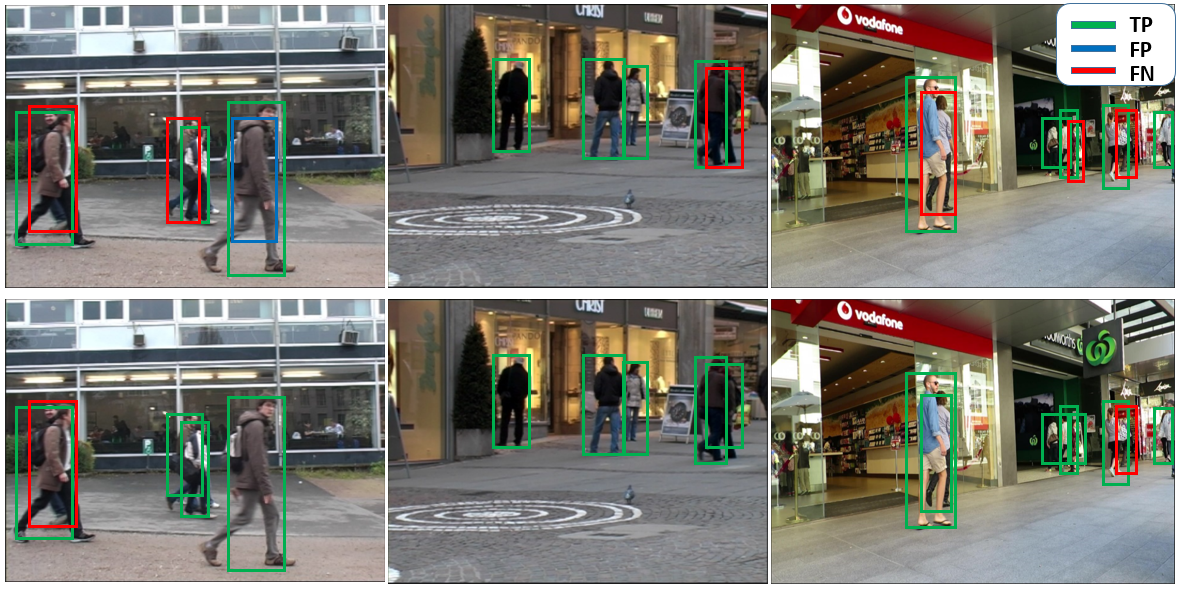}
%\vspace{-1em}
\caption{Visualized detections of GreedyNMS (the first row) and Pairwise-NMS (the second row). With green box, blue box, and red box to denote true positive, false positive and false negative, respectively. As can be been, compared with the detection results acquired from GreedyNMS, the detection results acquired from Pairwise-NMS dramatically reduce the false negatives (which leads to the higher recall), and meanwhile reduce the false positives slightly (which contributes to the precision).}

\label{fig5:vis_detection}
\end{figure*}
\end{center}

In Fig~\ref{fig4_nms_curves}, the MAP curves of best Soft-NMS\_L, best Soft-NMS\_G and Pairwise-NMS are given for the better illustration and easier comparison in visual. 

In conclusion, under the extremely crowded scenes, our proposed method can achieve a better balance between the recall and precision and is much robust than Soft-NMS.

%-------------------------------------------------------------------------

%\vspace{-0.5em}
\subsection{Qualitative Results}
\label{sec:qualitative_results}

In Fig
~\ref{fig5:vis_detection}, several visualized examples from MOT2015 dataset are given, the base detector is Faster-RCNN. The first row are the detection results we obtained using GreedyNMS, followed by the second row with results acquired by Pairwise-NMS. As can be seen from the legend at the top right, detections with blue-edge boxes are true positives (TP), detections with green-edge boxes are false positives (FP), and the detections with red-edge boxes are false negatives (FN).

In all of the three examples, due to the heavily overlapping between two or more person, GreedyNMS and Pairwise-NMS all failed in some cases and missing detect some targets (as denoted with the red boxes), however, compared with the results acquired from GreedyNMS, the detection obtained from Pairwise-NMS has less missing detections, which strongly prove the effectiveness of the proposed pairwise-relationship network for handling heavily occluded case. In addition, in the first image, as denoted by the blue box, Pairwise-NMS performs better in reducing false positives. To sum up, comparing to GreedyNMS, Pairwise-NMS effectively preventing the case of missing detections and discarding more false positives.

%----------------------------------------------------------------------------
\subsection{Ablation Study}
\label{sec4.4:ablation}

\subsubsection{Sampling Policy}
\label{appendix2:sample_pairs}

\noindent Due to the imbalance between similar pairs and dissimilar pairs, it is important to make a wise policy to achieve good accuracy and to ease the learning process of pairwise-relationship network. 

% \vspace{2mm}
% \noindent \textbf{Ratio between Positive and Negative Samples}

After a thorough investigation of datasets which contain heavy occlusions including TUD-Crossing and MOT15, we find the number of heavily occluded objects is relatively small compared with the total number of ground-truth bounding boxes. According to the analysis in Section~\ref{sec3.3: define_pairs}, the proposal pairs that contain two different objects are defined as dissimilar pairs, which should not be merged. All other cases are defined as similar pairs and should be merged. Based on our statistics on TUD-Crossing as can been seen in Table~\ref{table7:tud_pairs}, the ratio between dissimilar and similar pairs are more than 10. For the sake that the network can learn to recognize two objects, we randomly reduce the number of similar samples during training. Specifically, the ratio between dissimilar and similar pairs is kept with 1:3.

% \vspace{4mm}
% \noindent  \textbf{How to Choose Similar Pairs?}
Meanwhile, among the similar pairs, considering how many objects (zero or one) the two proposals contain, a natural way of sampling pairs is to form the similar pairs based on their ratios of different cases. Because correctly predicting the similar pairs will dramatically reduce the false positives and improve the precision, the sampling ratios will also affect network's ability to differentiate the dissimilar and the similar pairs.

\vspace{4mm}
\subsubsection{Different Network Structures}

\begin{center}
\begin{figure}[t]
\centering
\includegraphics[height=6.5cm]{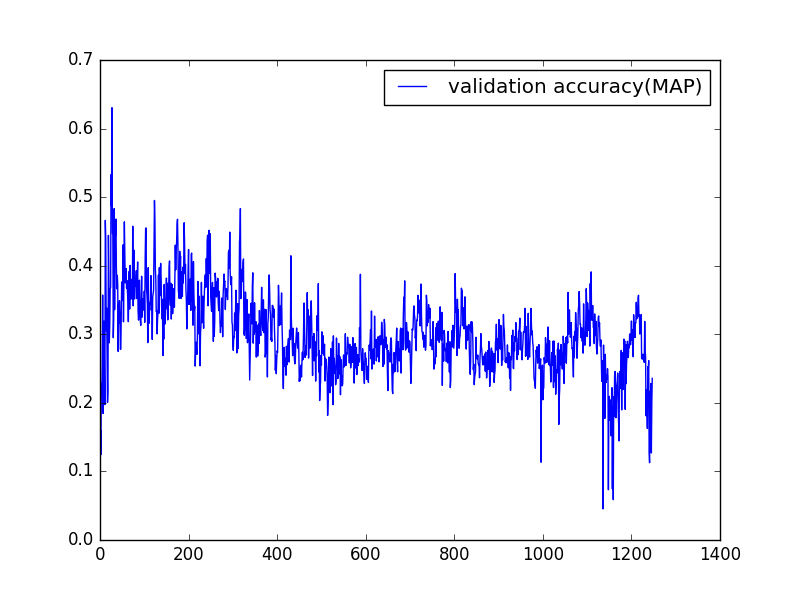}
%\vspace{-1em}
\caption{Validation accuracy of pairwise-relationship network using FC layer.}
\label{FC_curve}
\end{figure}
\end{center}

\begin{center}
\begin{figure}[t]
\centering
\includegraphics[height=6.5cm]{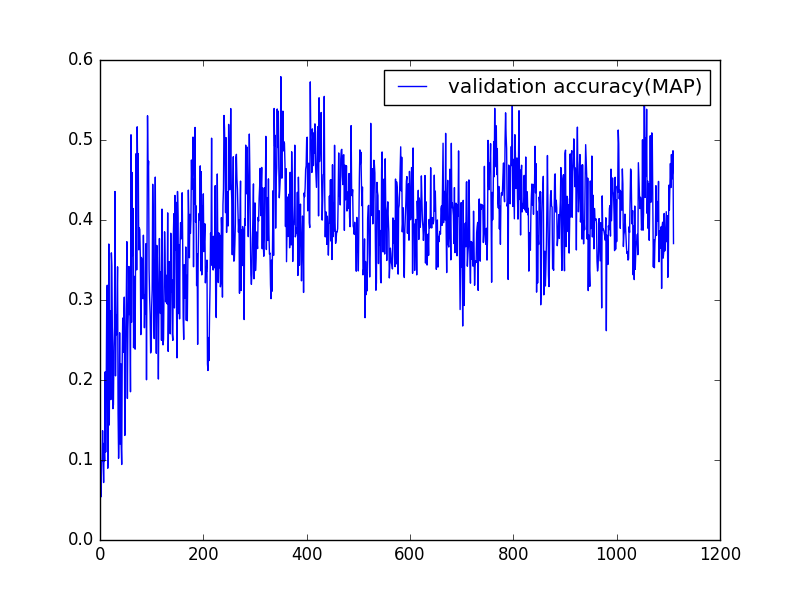}
%\vspace{-1em}
\caption{Validation accuracy of pairwise-relationship network using GAP layer}
\label{GAP_curve}
\end{figure}
\end{center}

In the experiments, we discover that, for the last layer, pairwise-relationship network with GAP layer has better performance than the one with FC layer. It may due to the reason that FC layer usually carries the position information, which in turn causes confusion to the network to distinguish the case5 from case6 in Table~\ref{table1:pair_relation}. Specifically, it is difficult for the pairwise-relationship network with FC layer to differentiate the case that two proposals contain one common object and the case that two proposals contain two different objects. 

% The difference between using FC layer and GAP layer is given in Supplementary (Section~\ref{GAPvsFC}) through the loss curves

In order to test the affects between using the global average pooling (GAP) layer and fully connected (FC) layer in the pairwise-relationship network. We use 80\% images of TUD-Crossing dataset to train the pairwise-relationship network with GAP layer and the pairwise-relationship network with FC layer from scratch, and use another 20\% images of TUD-Crossing as the validation set. The validation loss curves of the networks with GAP layer and FC layer are as shown in Fig~\ref{GAP_curve} and Fig~\ref{FC_curve}, respectively. In the two figures, X axis represents the number of training iteration, and Y axis is MAP (AP@0.5). As can be seen, the pairwise-relationship network with GAP layer has better performance and continue learning as the training goes. Which confirm our conjecture in the section of ablation Study, and for all of the experiments in this paper, the network with GAP layer is employed.  

\setlength{\tabcolsep}{3pt}
\begin{center}
\begin{table}[t]
\caption{Statistics of pairs on TUD-Crossing}
\label{table7:tud_pairs}
\scalebox{0.95}{
\begin{tabular}{llllll}
\hline\noalign{\smallskip}
proposals & IoU threshold &gt pairs &0 object pairs & 1 object pairs & 2 object pairs \\
\noalign{\smallskip}
\hline
\noalign{\smallskip}
300  & 0 & 537 & 980596 & 161320 & 7596\\
\noalign{\smallskip}
300 & 0.1 & 398 & 699854 & 133364 & 3061 \\
\noalign{\smallskip}
300 & 0.2 & 254 & 533469  & 74523 & 1582\\
\noalign{\smallskip}
300 & 0.3 & 150 & 391318 & 16309 & 930 \\
\noalign{\smallskip}
\hline
\end{tabular}
}
\end{table}
\end{center}
\setlength{\tabcolsep}{1.4pt}

%------------------------------------------------------------------------
%\vspace{-0.5em}
\section{Conclusions}
\label{sec5:conclusion}
We propose a Pairwise-NMS as the post-processing step for object detection. Feeding two overlapping proposals to a pairwise-relationship network can smartly tell the Pairwise-NMS if there are two objects or one/zero object contained. It can cure the drawback of GreedyNMS for ``suppressing all'' under heavy occlusions, and is more robust than Soft-NMS in crowded scenes. Moreover, Pairwise-NMS can consistently improve the performance and neatly couple with GreedyNMS. And can be as an ingredient to be integrated into learning-based detectors including Faster-RNN, DPM without losing the efficiency. We believe that our work is beneficial to instance segmentation and multi-object tracking in crowded scenes. 
In the future, there will be two research problems of our interests. One of the problems is to integrate Pairwise-NMS into learning based detectors such as Faster-RCNN, YOLO, SSD to perform joint training and inference, and the other is to explore a more general rule to handle a cluster of multiple objects at the same time.

\ifCLASSOPTIONcaptionsoff
  \newpage
\fi



%\bibliographystyle{IEEEtran}
% \end{thebibliography}

\bibliographystyle{IEEEtran}
\bibliography{IEEEabrv,reference}

\begin{thebibliography}{10}
\providecommand{\url}[1]{#1}
\csname url@samestyle\endcsname
\providecommand{\newblock}{\relax}
\providecommand{\bibinfo}[2]{#2}
\providecommand{\BIBentrySTDinterwordspacing}{\spaceskip=0pt\relax}
\providecommand{\BIBentryALTinterwordstretchfactor}{4}
\providecommand{\BIBentryALTinterwordspacing}{\spaceskip=\fontdimen2\font plus
\BIBentryALTinterwordstretchfactor\fontdimen3\font minus
  \fontdimen4\font\relax}
\providecommand{\BIBforeignlanguage}[2]{{%
\expandafter\ifx\csname l@#1\endcsname\relax
\typeout{** WARNING: IEEEtran.bst: No hyphenation pattern has been}%
\typeout{** loaded for the language `#1'. Using the pattern for}%
\typeout{** the default language instead.}%
\else
\language=\csname l@#1\endcsname
\fi
#2}}
\providecommand{\BIBdecl}{\relax}
\BIBdecl

\bibitem{Reference35}
A.~Arnab and P.~H. Torr, ``Pixelwise instance segmentation with a dynamically
  instantiated network,'' in \emph{CVPR}, 2017.

\bibitem{Reference11}
K.~He, G.~Gkioxari, P.~Dollar, and R.~Girshick, ``{Mask R-CNN},'' in
  \emph{ICCV}, 2017.

\bibitem{Reference36}
M.~B.~R. Urtasun, ``Deep watershed transform for instance segmentation,'' in
  \emph{CVPR}, 2017.

\bibitem{Reference17}
B.~Leibe, K.~Schindler, and L.~V. Gool, ``Coupled detection and trajectory
  estimation for multi-object tracking,'' in \emph{ICCV}, 2007.

\bibitem{Reference29}
S.~Tang, M.~Andriluka, A.~Milan, K.~Schindler, S.~Roth, and B.~Schiele,
  ``Learning people detectors for tracking in crowded scenes,'' in \emph{ICCV},
  2013.

\bibitem{Reference18}
A.~R. Zamir, A.~Dehghan, and M.~Shah, ``{GMCP-Tracker:} global multi-object
  tracking using generalized minimum clique graphs,'' \emph{ECCV}, 2012.

\bibitem{Reference30}
M.~A. Sadeghi and A.~Farhadi, ``Recognition using visual phrases,'' in
  \emph{CVPR}, 2011.

\bibitem{Reference31}
W.~Ouyang and X.~Wang, ``Single-pedestrian detection aided by multi-pedestrian
  detection,'' in \emph{CVPR}, 2013.

\bibitem{Reference32}
M.~Rodriguez, I.~Laptev, and J.~Sivic, ``Density-aware person detection and
  tracking in crowds,'' in \emph{ICCV}, 2011.

\bibitem{Reference2}
Y.~LeCun, L.~Bottou, Y.~Bengio, and P.~Haffner, ``Gradient-based learning
  applied to document recognition,'' \emph{IJCV}, vol.~86, no.~11, pp.
  2278--2324, 1998.

\bibitem{Reference3}
A.~Krizhevsky, I.~Sutskever, and G.~E. Hinton, ``Imagenet classification with
  deep convolutional neural networks,'' in \emph{NIPS}, 2012.

\bibitem{Reference4}
O.~Russakovsky, J.~Deng, H.~Su, J.~Krause, S.~Satheesh, S.~Ma, Z.~Huang,
  A.~Karpathy, A.~Khosla, M.~Bernstein, A.~C. Berg, and L.~Fei-Fei, ``Imagenet
  large scale visual recognition challenge,'' \emph{IJCV}, vol. 115, no.~3, pp.
  211--252, 2015.

\bibitem{Reference13}
K.~He, X.~Zhang, S.~Ren, and J.~Sun, ``Deep residual learning for image
  recognition,'' in \emph{CVPR}, 2016.

\bibitem{Reference5}
R.~Girshick, J.~Donahue, T.~Darrell, and J.~Malik, ``Rich feature hierarchies
  for accurate object detection and semantic segmentation,'' in \emph{CVPR},
  2012.

\bibitem{Reference8}
R.~Girshick, ``{Fast R-CNN},'' in \emph{ICCV}, 2015.

\bibitem{Reference9}
S.~Ren, K.~He, R.~Girshick, and J.~Sun, ``{Faster R-CNN}: Towards real-time
  object detection with region proposal networks,'' in \emph{NIPS}, 2015.

\bibitem{Reference10}
K.~He, X.~Zhang, S.~Ren, and J.~Sun, ``Spatial pyramid pooling in deep
  convolutional networks for visual recognition,'' in \emph{ECCV}, 2014.

\bibitem{Reference14}
T.-Y. Lin, P.~Dollar, R.~Girshick, K.~He, B.~Hariharan, and S.~Belongie,
  ``Feature pyramid networks for object detection,'' in \emph{CVPR}, 2017.

\bibitem{Reference21}
P.~Felzenszwal, R.~Girshick, and D.~McAllester, ``Object detection with
  discriminatively trained part-based models,'' \emph{TPAMI}, vol.~32, no.~9,
  pp. 1627--1645, 2010.

\bibitem{Reference12}
T.-Y. Lin, M.~Maire, S.~Belongie, L.~Bourdev, R.~Girshick, J.~Hays, P.~Perona,
  D.~Ramanan, C.~L. Zitnick, and P.~Dollar, ``{Microsoft COCO:} common objects
  in context,'' in \emph{ECCV}, 2014.

\bibitem{Reference19}
M.~Everingham, S.~M.~A. Eslami, L.~V. Gool, C.~K.~I. Williams, J.~Winn, and
  A.~Zisserman, ``The pascal visual object classes challenge: A
  retrospective,'' \emph{IJCV}, vol. 111, no.~1, pp. 98--136, 2015.

\bibitem{Reference26}
N.~Bodla, B.~Singh, R.~Chellappa, and L.~S. Davis, ``{Soft-NMS}-- improving
  object detection with one line of code,'' in \emph{ICCV}, 2017.

\bibitem{Reference41}
S.~H. Rezatofighi, V.~Kuma, A.~Milan, E.~Abbasnejad, A.~Dick, and I.~Reid,
  ``{DeepSetNet:} predicting sets with deep neural networks,'' in \emph{ICCV},
  2017.

\bibitem{Reference20}
H.~Pirsiavash, D.~Ramanan, and C.~C. Fowlkes, ``Globally optimal greedy
  algorithms for tracking a variable number of objects,'' in \emph{CVPR}, 2011.

\bibitem{Reference22}
T.~T. Pham, S.~H. Rezatofighi, I.~Reid, and T.-J. Chin, ``Efficient point
  process inference for large-scale object detection,'' in \emph{CVPR}, 2016.

\bibitem{Reference23}
O.~Barinova, V.~Lempitsky, and P.~Kholi, ``On detection of multiple object
  instances using hough transforms,'' \emph{TPAMI}, vol.~34, no.~9, pp.
  1773--1784, 2012.

\bibitem{Reference24}
A.~B. Chan, Z.-S.~J. Liang, and N.~Vasconcelos, ``Privacy preserv- ing crowd
  monitoring: Counting people without people mod- els or tracking,'' in
  \emph{CVPR}, 2008.

\bibitem{Reference25}
R.~Rothe, M.~Guillaumin, and L.~V. Gool, ``Non-maximum suppression for object
  detection by passing messages between windows,'' in \emph{ACCV}, 2014.

\bibitem{Reference27}
J.~Hosang, R.~Benenson, and B.~Schiele, ``A convnet for non-maximum
  suppression,'' in \emph{GCPR}, 2016.

\bibitem{Reference28}
------, ``Learning non-maximum suppression,'' in \emph{CVPR}, 2017.

\bibitem{Reference7}
K.~Simonyan and A.~Zisserman, ``Very deep convolutional networks for
  large-scale image recognition,'' in \emph{arXiv preprint arXiv:1409.1556},
  2014.

\bibitem{szegedy2016rethinking}
C.~Szegedy, V.~Vanhoucke, S.~Ioffe, J.~Shlens, and Z.~Wojna, ``Rethinking the
  inception architecture for computer vision,'' in \emph{Proceedings of the
  IEEE conference on computer vision and pattern recognition}, 2016, pp.
  2818--2826.

\bibitem{Reference37}
M.~Lin, Q.~Chen, and S.~Yan, ``Network in network,'' in \emph{arXiv preprint
  arXiv:1312.4400}, 2013.

\bibitem{Reference40}
S.~Ioffe and C.~Szegedy, ``{Batch Normalization:} accelerating deep network
  training by reducing internal covariate shift,'' in \emph{arXiv preprint
  arXiv:1502.03167}, 2015.

\bibitem{Reference38}
L.~Leal-Taixe, A.~Milan, I.~Reid, S.~Roth, and K.~Schindler, ``{MOTChallenge
  2015:} towards a benchmark for multi-target tracking,'' in \emph{arXiv
  preprint arXiv:1504.01942}, 2015.

\bibitem{Reference39}
M.~Andriluka, S.~Roth, and B.~Schiele, ``People-tracking-by-detection and
  people-detection-by-tracking,'' in \emph{CVPR}, 2008.

\bibitem{Reference42}
A.~Ellis and J.~M. Ferryman, ``{PETS2010 and PETS2009} evaluation of results
  using individual ground truthed single views,'' in \emph{AVSS}, 2010.

\end{thebibliography}

\end{document}